\documentclass[sigconf,natbib=true,anonymous=false]{acmart}

\usepackage{tabularx}
\usepackage{bigstrut}
\usepackage{booktabs}
\usepackage{multirow}
\usepackage{microtype}
\usepackage{graphicx}
\usepackage{subfig}
\usepackage{overpic}
\usepackage{balance}
\usepackage{amsthm}
\usepackage{array}
\usepackage{clrscode}
\usepackage{multicol}
\usepackage{float}
\usepackage{hyperref} 
\usepackage{newfloat}
\usepackage{color}
\usepackage{xcolor}
\usepackage{amsopn}
\usepackage{mathrsfs}
\usepackage{mathtools}
\usepackage{amsmath}
\usepackage{arydshln}
\usepackage{blkarray}
\usepackage{enumerate}
\usepackage{courier}
\usepackage{rotating}
\usepackage{bm}
\usepackage{wrapfig}
\usepackage{ragged2e}
\usepackage[misc]{ifsym}
\usepackage[most]{tcolorbox}
\usepackage{threeparttable}
\usepackage{makecell}
\usepackage{adjustbox} 
\usepackage{enumitem}
\usepackage{hyperref}
\usepackage{url}
\usepackage{xspace}
\usepackage{comment}
\usepackage{changepage}
\usepackage{algorithm}
\usepackage{algorithmic}
\usepackage{calc}

\definecolor{codegreen}{rgb}{0,0.3,0.6}
\definecolor{codegray}{rgb}{0.5,0.5,0.5}

\newcommand{\paratitle}[1]{\vspace{1.5ex}\noindent\textbf{#1}}
\newcommand{\wrt}{w.r.t.\xspace}

\newcommand{\ignore}[1]{}

\AtBeginDocument{%
  \providecommand\BibTeX{{%
    \normalfont B\kern-0.5em{\scshape i\kern-0.25em b}\kern-0.8em\TeX}}}

\setcopyright{acmlicensed}
\copyrightyear{2018}
\acmYear{2018}
\acmDOI{XXXXXXX.XXXXXXX}

\acmConference[Conference acronym 'XX]{Make sure to enter the correct
  conference title from your rights confirmation emai}{June 03--05,
  2018}{Woodstock, NY}
\acmISBN{978-1-4503-XXXX-X/18/06}

\begin{document}


\title{
RecNet: Self-Evolving Preference Propagation for Agentic Recommender Systems}

\author{Bingqian Li$^*$}
\affiliation{%
    \institution{GSAI, Renmin University of China}
    \city{Beijing}
    \country{China}
}
\email{fortilinger@ruc.edu.cn}

\author{Xiaolei Wang$^{*}$}
\affiliation{%
    \institution{GSAI, Renmin University of China}
    \city{Beijing}
    \country{China}
}
\email{xiaoleiwang@ruc.edu.cn}

\author{Junyi Li}
\affiliation{%
    \institution{Department of Data Science, City University of Hong Kong}
    \city{Hong Kong}
    \country{China}
}
\email{junyili@cityu.edu.hk}

\author{Weitao Li}
\affiliation{
    \institution{Meituan}
    \city{Beijing}
    \country{China}
}
\email{liweitao05@meituan.com}

\author{Long Zhang}
\affiliation{
    \institution{Meituan}
    \city{Beijing}
    \country{China}
}
\email{zhanglong40@meituan.com}

\author{Sheng Chen}
\affiliation{
    \institution{Meituan}
    \city{Beijing}
    \country{China}
}
\email{chensheng19@meituan.com}

\author{Wayne Xin Zhao\textsuperscript{\Letter}}
\orcid{0000-0002-8333-6196}
\affiliation{
    \institution{GSAI, Renmin University of China}
    \city{Beijing}
    \country{China}
}
\email{batmanfly@gmail.com}

\author{Ji-Rong Wen}
\orcid{0000-0002-9777-9676}
\affiliation{
    \institution{
    GSAI, Renmin University of China}
    \city{Beijing}
    \country{China}
}
\email{jrwen@ruc.edu.cn}

\thanks{$^*$ Equal contribution.}
\thanks{\Letter \ Corresponding author.}
\thanks{GSAI is the abbreviation of Gaoling School for Artificial Intelligence.}

\renewcommand{\shortauthors}{Bingqian Li, et al.}

\begin{abstract}
Agentic recommender systems leverage Large Language Models (LLMs) to model complex user behaviors and support personalized decision-making. 
However, existing methods primarily model preference changes based on explicit user–item interactions, which are sparse, noisy, and unable to reflect the real-time, mutual influences among users and items.
To address these limitations, we propose RecNet, a self-evolving preference propagation framework that proactively propagates real-time preference updates across related users and items. The RecNet consists of two complementary phases. In the forward phase, the centralized preference routing mechanism leverages router agents  to integrate preference updates and dynamically propagate them to the most relevant agents. To ensure accurate and personalized integration of propagated preferences, we further introduce personalized preference reception mechanism, which combines a message buffer for temporary caching and an optimizable, rule-based filter memory to guide selective preference assimilation based on past experience and interests. In the backward phase, the feedback-driven propagation optimization mechanism simulates a multi-agent reinforcement learning framework, using LLMs for credit assignment, gradient analysis, and module-level optimization, enabling continuous self-evolution of propagation strategies. 
Extensive experiments on various scenarios demonstrate the effectiveness of RecNet in modeling preference propagation for recommender systems.

\end{abstract}



\maketitle

\section{Introduction}
\label{sec:intro}

Recently, Large Language Models (LLMs) have demonstrated remarkable capabilities in text understanding and generation~\cite{zhao2023survey, wang2024survey}, and have been increasingly leveraged as autonomous agents in diverse applications~\cite{hu2025agentgen, recmind}. Through memory mechanisms and multi-turn interactions, LLM-based agents exhibit strong potential for recommender systems by simulating user decision-making processes and enabling personalized interactions~\cite{recmind, llmrec}.

\begin{figure}[]
\centering
\includegraphics[width=0.94\linewidth]{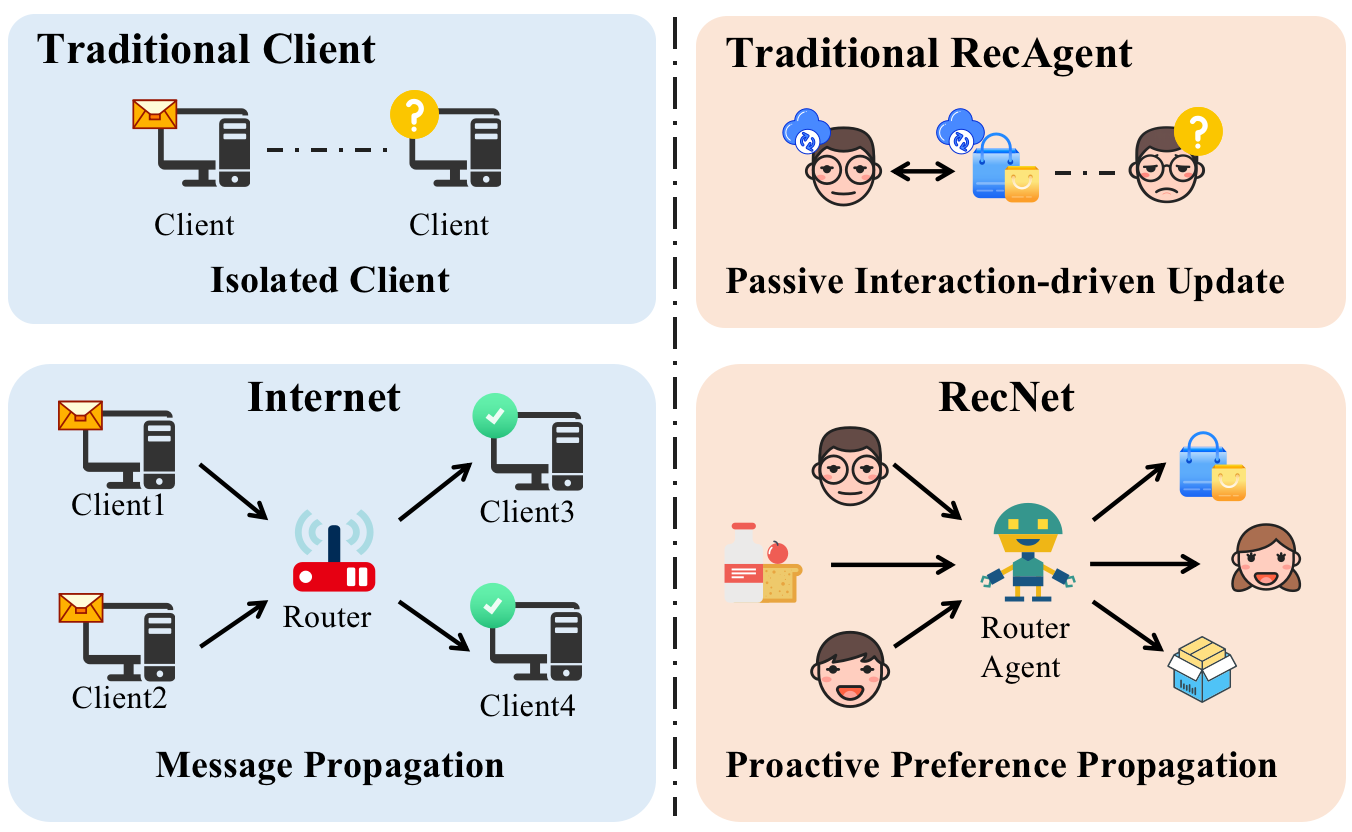}
\caption{An Analogy between Internet Routing Mechanism and RecNet Preference Propagation Mechanism.}
\label{fig:intro}
\end{figure}

Existing LLM-based recommendation agents~\cite{agent4rec, recmind} primarily simulate user behaviors to better understand users and model their evolving preference profiles. Beyond user-side modeling, several works ~\cite{agentcf, kgla, agentcf++} further consider the item-side modeling, simulating how an item’s representation evolves as it attracts different types of users over time.
Despite these advances, existing approaches fundamentally rely on {explicit user-item interactions} as the sole trigger for preference updates. Specifically, user or item profiles are updated only when explicit user-item interactions occur, such as clicking, rating, or purchasing.
However, in real-world recommendation scenarios, users and items constitute a dynamic and interconnected network shaped by diverse relationships. Beyond direct user-item
interactions, preference updates in one user or item profile may propagate through implicit relations (e.g.,  social connections, content co-engagement, or shared communities), thereby inducing correlated preference changes in similar users or items even without direct interactions.
As a result, purely interaction-driven methods update profiles passively based on sparse and noisy interaction data, and therefore lag behind in modeling and capturing dynamic preference changes for users and items.


These limitations motivate a shift from passive, interaction triggered preference updates toward a {proactive preference propagation} paradigm, in which the updated preference information is proactively propagated to highly-related users and items, so as to more accurately model the evolving preferences of them.
A naïve propagation strategy is to use full preference profiles to identify related users/items, and then directly propagate these profiles to them as the transmitted information. However, such point-to-point propagation fails to address two fundamental challenges:~(1) \emph{Scalability and relevance filtering}: In large-scale systems with frequent preference updates, propagation mechanisms must selectively transmit only essential information within updated profiles to the most relevant user and item communities.
~(2) \emph{Error control and adaptability}: Preference propagation may introduce noise or misalignment, requiring the propagation strategy to be continuously optimized based on interacation feedback.

To address these challenges, we introduce \emph{router agents} as intermediaries between user and item agents, enabling agentic and adaptive preference propagation.
This design is inspired by computer networks, where routers coordinate data transmission among clients, as shown in Figure~\ref{fig:intro}.
Therefore, we consider users and items as client agents and build a client-router-client preference propagation pipeline. Each router agent is designed to coordinate a community of similar client agents, integrating incoming preference information from updated clients and dynamically routing them to the most relevant downstream agents. 
The merits of designing and agentifying router modules for preference propagation are twofold:
~(1) Router agents serve as intermediate filters and integrators between updated and propagated users / items, effectively filtering irrelevant information and ensuring precise and efficient preference propagation.
~(2) Based on agentic capabilities, router agents can continuously interpret feedback signals during test-time propagation and autonomously adapt their propagation strategies, enabling the self-evolving process of preference propagation.

To this end, we propose \textbf{RecNet}, a self-evolving preference propagation framework for recommender systems that introduces router agents to proactively propagate real-time preference updates across related users and items.
RecNet departs from conventional interaction-driven paradigms by enabling preference information to flow through implicit relational structures, allowing the system to capture correlated and timely preference dynamics.
Specifically, RecNet makes three key technical contributions: (1) \textbf{Centralized preference routing}, where router agents act as intermediaries to perceive, integrate, and dynamically route preference updates to the most relevant user and item communities. (2) \textbf{Personalized preference reception}, which allows each user and item agent to selectively assimilate propagated preferences through a message buffer and an optimizable, rule-based filter memory, preserving personalization during propagation. (3) \textbf{Feedback-driven propagation optimization}, which leverages interaction feedback and textual backpropagation to continuously refine routing strategies, enabling RecNet to self-evolve under dynamic environments.
Based on these components, RecNet operates in two phases: forward preference propagation that proactively disseminates preference updates, and backward preference optimization that continuously improves propagation behaviors through feedback signals.

The main contributions of this paper are as follows:

$\bullet$ We propose RecNet, a self-evolving preference propagation framework for recommender systems. By introducing router agents as intermediaries, RecNet enables centralized preference routing and personalized preference reception to proactively integrate and propagate preference updates across users and items.

$\bullet$ We develop a feedback-driven propagation optimization mechanism that leverages a textual optimization strategy to adapt propagation behaviors based on interaction feedback, enabling the continuous self-evolution of preference propagation strategies.

$\bullet$ Empirical evaluations on three datasets and various scenarios demonstrate the effectiveness of RecNet.
\section{Methodology}
\label{sec:method}

\subsection{Overview of Our Approach}

\paratitle{Problem Formulation.}
In traditional recommender systems, given a set of users $\mathcal{U}$, a set of items $\mathcal{I}$, and their interaction records, the goal is to model user preferences from historical interactions and recommend items that best match these preferences.
Building upon this foundational understanding, in agentic recommendation framework, each user $u$ and item $i$ is associated with a textual preference profile, denoted as $P_u$ and $P_i$, respectively. These profiles are generated and maintained by large language models (LLMs), where $P_u$ captures the user's evolving preferences and $P_i$ represents the item's semantic characteristics. The task is to refine $P_u$ and $P_i$ in recommender systems, such that for a candidate item set $\mathcal{I}_c$, the system can generate a ranking list that accurately prioritizes items aligned with the user’s preferences.

\begin{figure*}[h]
\centering
\captionsetup{skip=5pt}
\includegraphics[width=0.94\linewidth]{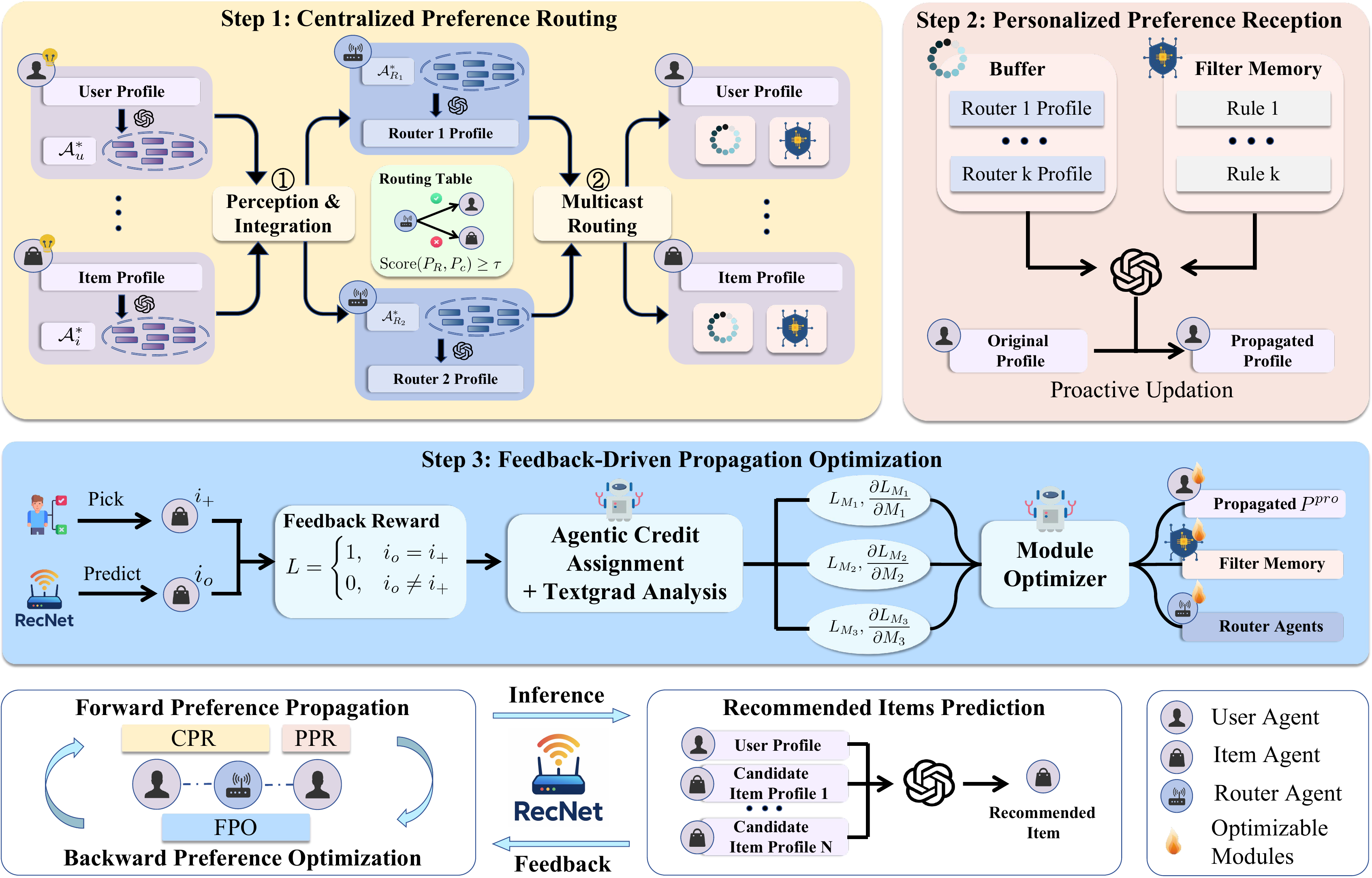}
\caption{The overall framework of RecNet.
}
\label{fig:model}
\vspace{-10pt}
\end{figure*}

\paratitle{Overview of RecNet.}
To dynamically capture the mutual influences among communities of users and items in recommender systems, an agentic recommender should not only model passive interaction-driven updates in preference profiles, but also explicitly incorporate proactive preference propagation. Inspired by the client–router–client architecture in computer networks, 
we propose \textbf{RecNet}, a self-evolving preference propagation framework, which comprises a forward propagation pipeline to propagate real-time updated preference information, and a backward optimization pipeline to dynamically refine propagation strategies based on environmental feedback.
In the \emph{forward propagation} pipeline, we introduce a centralized preference routing mechanism, where router agents act as intermediaries between user and item agents. These router agents integrate and dynamically route updated preference information to the most relevant users and items, enabling accurate and efficient preference propagation (Section~\ref{sec:central}).
To preserve personalization during propagation, we further design a personalized preference reception mechanism, which allows each user and item agent to selectively assimilate propagated preferences through a message buffer and an optimizable, rule-based filter memory~(Section~\ref{sec:personal}). 
In the \emph{backward optimization} pipeline, we propose a feedback-driven propagation optimization mechanism that leverages interaction feedback and textual signals to guide the refinement of propagation-related modules~(Section~\ref{sec:evolving}). The overall framework of RecNet is shown in Figure~\ref{fig:model}.

\subsection{Centralized Preference Routing}
\label{sec:central}
To effectively model preference propagation in recommender systems, several key requirements must be satisfied.
First, the propagation process should be precise and selective. The updated preference information should be propagated only to users or items that are highly relevant, so as to prevent weakly correlated or redundant signals from affecting profile updates. This requires the system to carefully manage both the content of the propagated updates and the targets to which they are sent.
Second, preference propagation often exhibits a community-level broadcasting effect, where updates can jointly influence groups of users or items with similar interests. The propagation process should leverage these community-level effects to mitigate the impact of isolated or noisy individual behaviors, propagating signals that accurately capture the shared interests within the community.
Considering these two attributes of preference propagation, direct profile-level point-to-point propagation becomes both unreliable and computationally expensive, leading to suboptimal performance, as shown in Table~\ref{tab:propagation}.

To address these challenges, an central propagation manager is required to aggregate, filter, and redistribute preference information in a more accurate and efficient manner. 
Inspired by the client-router-client architecture in computer networks, where routers coordinate communication among groups of connected clients by maintaining routing states and dynamically selecting transmission paths, we introduce router agents as intermediaries between user and item agents for preference propagation. Each router agent serves as a community-level coordinator, whose profile represents the shared attributes and common preference patterns of a group of semantically related users and items.
By operating at the community level, router agents enable structured aggregation and selective dissemination of preference updates, thereby improving both the effectiveness and efficiency of preference propagation.
In the following, we will describe the centralized preference routing mechanism in detail, including (1) initialization, (2) client-to-router preference perception and integration, and (3) router-to-client multi-cast preference routing.

\subsubsection{Initialization of Centralized Preference Routing}
The first step of the centralized preference routing module is to enable router agents to accurately perceive and integrate newly updated preference information from client agents, including both user and item agents.
Since client preference profiles are fine-grained and frequently updated, directly propagating full textual profiles is inefficient and ineffective for capturing evolving preference dynamics. Therefore, we transform each client profile into a more granular representation that better reflects its dynamic preference state. Specifically, each client agent $c$ is represented from two complementary perspectives: its textual preference profile $P_c$, and a set of fine-grained preference attributes $\mathcal{A}_c$. 
To extract and structure these attributes, we leverage the analytical capabilities of LLMs to identify and summarize relevant keywords and concepts from $P_c$, forming a granular representation of the client’s updated preferences by using $\text{Prompt}_{extract}$:
\begin{equation}
\mathcal{A}_c = \text{Prompt}_{extract}(P_c).
\end{equation}
At the initialization stage of RecNet, the extracted attribute sets from all client agents are clustered into $K$ groups, each corresponding to a router agent in the set $\mathcal{R}=\{R_1, R_2, \dots, R_K\}$. 
This clustering process allows each router agent to form a community-level representation that captures the shared preference attributes of a group of semantically related users and items. These initialized router agents serve as the foundation for subsequent preference propagation, supporting the dynamic updating of router profiles and guiding the routing of newly updated preference information throughout the network.

\subsubsection{Client-to-Router Preference Perception and Integration}
At each time step $t$, when certain clients’ preferences or characteristics change, the fine-grained dynamics are captured in the updates to $\mathcal{A}_c$, and the updated attributes are aggregated into a list of newly updated attributes, $\mathcal{A}_t$. To efficiently route $\mathcal{A}_t$ to the most relevant user and item communities, each attribute $a \in \mathcal{A}_t$ is treated as a data packet and broadcast to the set of router agents $\mathcal{R}t = {R_1, R_2, \dots, R_{K_t}}$ for preference propagation, where $K_t$ denotes the number of routers at time step $t$ (Note that both the number and profiles of these router agents evolve dynamically based on interaction feedback; see Section~\ref{sec:evolving}).

Upon receiving the updated attributes $\mathcal{A}_t$, the router agents perform a two-step preference perception and integration process.
First, to ensure both semantic relevance and clustering efficiency, we employ a strong LLM encoder to transform each attribute $a \in \mathcal{A}_t$ into its embedding $e_a$, and each router profile $P_R$ into $e_R$.
Next, each attribute embedding $e_a$ is routed to the most relevant router by computing its similarity with all router embeddings $\{e_R\}_{R \in \mathcal{R}_t}$. 
Specifically, the target router $R^*$ for attribute $a$ is determined as:
\begin{equation}
R^* = \arg\max_{R \in \mathcal{R}_t} \; \text{sim}(e_a, e_R),
\end{equation}
where $\text{sim}(\cdot, \cdot)$ denotes cosine similarity. After that, with LLMs' strong abilities in summarization and generation, we prompt LLMs to summarize the profile of each router agent with newly added attributes, forming a new router profile for each router:
\begin{align}
P_R^* = \text{Prompt}_{summarize}(\{a\}, P_R), a \in \mathcal{A}_R,
\end{align}
in which $\mathcal{A}_R$ denotes the set of newly clustered attributes for $R$, and $\text{Prompt}_{summarize}$ is the summarization prompt.
After perceiving and integrating updated preference information, each router agent captures the preference dynamics relevant to its community. The updated router profile can be used to identify and propagate information to clients within the community.

\subsubsection{Router-to-Client Multicast Preference Routing}
After preference perception and integration, the next step is to dynamically construct the connections between router agents and client agents, so that they can route their preference information to highly-related users or items. 
Considering the large amounts of users and items, we design a similarity score that can be calculated efficiently to form a similarity matrix, denoted as a routing table, between updated router agents and client agents:
\begin{align}
\text{Score}(R, c) = \text{sim}(e_{P_R}, e_{P_c}) \times \mathbb{I}\{\mathcal{A}_R \cap \mathcal{A}_c \neq \emptyset\}.
\end{align}
Then, we select connections with high similarity scores above a threshold $\tau$ to form the propagation routing paths, propagating the router's updated profile as the messages to related users and items.

The design of router agents with the above propagation mechanisms offers two main advantages.
(1) Precision in propagation: By capturing fine-grained preference updates and dynamically selecting target agents based on the current community preference~(router profile), router agents can propagate the most precise and concise interest information to the most relevant users and items.
(2) Efficiency in dissemination: Instead of requiring clients to engage in costly point-to-point communication to determine what and whom to propagate, a limited number of router agents are enough to dynamically manage specific user and item communities through their router profiles. This enables relevant users / items to receive preference updates with much lower computational and communication overhead.

\subsection{Personalized Preference Reception}
\label{sec:personal}
With the Preference Routing mechanism, every client agent can receive updated and condensed preference information from similar users or items in a timely and dynamic manner. However, directly fusing the current profile with these propagated preference information will lead to two problems. 
Firstly, different agents should selectively integrate incoming preferences based on their own profiles and past experiences. 
Secondly, the interaction frequency varies across users and items. For those with infrequent interactions, if propagated information is indiscriminately and frequently merged without interaction-based validation, their profiles may easily deviate from true preferences. 
To address this, we introduce a buffer module and a filter memory module to each client agent for effective and personalized integration of propagated information.

\subsubsection{Queue-Based Buffer for Preference Information Caching}
Continuously merging propagated preferences based solely on propagation frequency, rather than personalized interaction frequency, can quickly distort a profile, especially for users or items with infrequent interactions.
To address this, we maintain a queue-based buffer for each user or item to cache incoming propagated preferences. These cached preferences are fused into the profile either when the profile is accessed or just before the next interaction, ensuring the profile represents up-to-date preferences.
Using a last-in, first-out strategy with a limited buffer capacity ensures that the profile always incorporates the most recent pieces of preference changes, keeping it timely and relevant to evolving interests.

\subsubsection{Rule-based Filter Memory for Preference Information Integration}
Different users or items may personalize how they integrate propagated preference information, even when receiving the same updates. Such personalization depends not only on current preferences, but also on how each client historically responds to different types of propagated signals. To support this behavior, we introduce a rule-based filter memory, which serves as an integration control mechanism and stores a set of textual integration rules that guide how propagated preference information should be selectively incorporated into the current client profile. These rules encode integration preferences (e.g., ``prioritize preference updates related to jazz and R\&B music''), and are dynamically updated based on interaction feedback. Importantly, the filter memory does not represent user preferences themselves; instead, it captures how preference information should be filtered and weighted during profile integration. During preference reception, the filter memory is injected as contextual guidance into the prompt used by the LLM to merge propagated information with the existing profile.

With personalized preference reception mechanism, the final profile $P_c^{pro}$ for client $c$ after integration can be written as follows:
\begin{align}
P_c^{pro} = \text{Prompt}_{merge}(\{R_c\}, F_c, P_c),
\end{align}
in which $\{R_c\}$ denotes the set of propagated router profiles within the buffer $B_c$, $F_c$ denotes the personalized filter memory, $P_c$ denotes the original client profile, and $\text{Prompt}_{merge}$ denotes the prompt for formulating the propagated profile.

\subsection{Feedback-driven Propagation Optimization}
\label{sec:evolving}

\begin{figure}[]
\centering
\captionsetup{skip=5pt}
\includegraphics[width=0.87\linewidth]{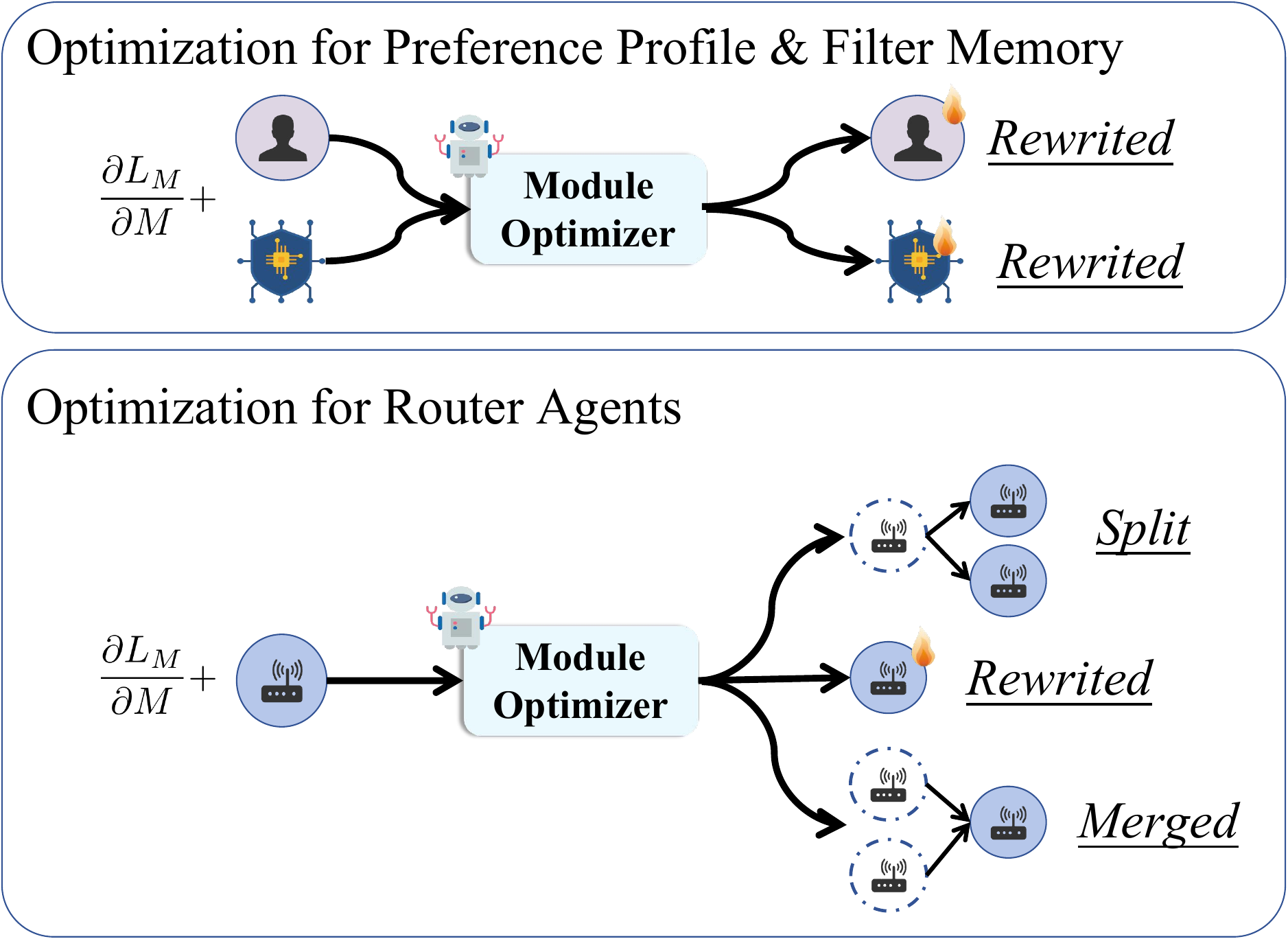}
\caption{The operation of module optimizer considering different optimized modules.}
\label{fig:optimization}
\vspace{-10pt}
\end{figure}

With centralized preference routing and personalized preference reception, we construct a feed-forward pipeline that propagates updated preference information to similar users and items both effectively and efficiently. However, existing preference propagation systems often rely on heuristic designs or human-centered initialization, and thus lack adaptability to dynamically evolving recommendation environments.
For example, with a fixed number of router agents, each router is forced to absorb increasingly diverse interests as user preferences expand, leading to redundant and less discriminative propagation. Moreover, non-learnable propagation strategies may introduce systematic bias and cause performance drift over time.

To address these limitations, we leverage interaction feedback not only to optimize preference profiles, but also to dynamically refine the entire propagation mechanism in the feed-forward pipeline, including individual filter memories, router profiles, and even the number of router agents.

We begin by collecting feedback on the quality of the propagated profiles. 
As users and items are symmetrically modeled as client agents in RecNet, we use a user-side interaction to illustrate the feedback construction process.
After propagation, the user profile is updated to $P_u^{pro}$. 
We then form a candidate set containing the positive item $i^+$ and a randomly sampled negative item $i^-$. 
Given the updated user profile and the updated profiles of these two items, we prompt the LLM with $\text{Prompt}_{predict}$ to predict which item the user is more likely to interact with.

\begin{align}
i_o = \text{Prompt}_{predict}(P_u^{pro}, P_{i+}^{pro}, P_{i-}^{pro}).
\end{align}
Then, we can assess the accuracy of the user's profile modeling by checking whether the selected item $i_o$ matches the truly interacted item. The feedback reward $L$ is defined as follows: 
\begin{align}
L = 
\begin{cases}
  1, & i_o = i_+ \\
  0, & i_o \neq i_+
\end{cases}
\end{align}
 
Given this reward signal, a natural question is how to use it to optimize the propagation process. Traditional reinforcement learning frameworks typically rely on scalar reward signals and parameterized, gradient-based updates, which are computationally expensive and difficult to support in dynamically evolving online recommendation settings. 
Also, in RecNet, the key components—including textual filter memories, router profiles, and even the number of router agents—are inherently language-based and non-parametric, making them unsuitable for direct parameter updates. 
To address both challenges, we adopt a textual optimization paradigm that simulates a multi-agent reinforcement learning framework with two stages: (i) credit assignment and gradient analysis, and (ii) module optimization.

\paratitle{Credit Assignment and Gradient Analysis.} 
First, to transform coarse-grained outcome feedback into fine-grained, module-level credit assignments for RecNet, we prompt the LLM with $\text{Prompt}_{gradient}$ to evaluate the contribution of each module $M$ involved in generating $P_u^{pro}$ based on the feedback reward $L$. 
\begin{align}
\label{equa:gradient}
L_M, \frac{\partial L_M}{\partial M} = \text{Prompt}_{gradient}(M, L).
\end{align}

Analogous to reward assignment and gradient computation in multi-agent reinforcement learning, here we prompt the LLM to output a "textual reward" $L_M$ that evaluates the module's impact on the final decision, and a "textual gradient" $\frac{\partial L_M}{\partial M}$ that contains executable suggestions for revising or optimizing the module. This approach enables module-level optimization in a language-aware, non-parametric manner, transforming scalar feedback signals into interpretable and actionable instructions that RecNet’s propagation modules can understand and execute.


\paratitle{Module Optimization.}
Similar to the function of optimizer in parameterized optimization method, which directly optimizes the parameter based on the provided gradients, we update each module in RecNet based on the textual gradient provided using $\text{Prompt}_{optimizer}$ as follows:

\begin{align}
M^* = \text{Prompt}_{optimizer}(M, \frac{\partial L}{\partial M}).
\end{align}
Specifically, different modules $M$ in RecNet require distinct optimization operations. For user/item profiles $P$ and filter memories $F$, we prompt the LLM to rewrite their contents according to the provided textual gradients. 
For centralized router agents, multiple client agents may produce textual gradients with respect to the same router. Therefore, we aggregate these signals and prompt the LLM to autonomously decide whether to split, merge, or rewrite each router agent, enabling dynamic and self-evolving adaptation of the routing strategy. 
The module-specific optimization operations are illustrated in Figure~\ref{fig:optimization}.

With feedback-driven propagation optimization, the network achieves life-long continuous training, where feedback from interactions not only refreshes user or item's representations but also drives the self-evolution of propagation strategies. This enables the system to remain adaptive and robust under dynamically changing recommendation environments, without relying on heuristic human-designed rules or static assumptions.

\subsection{The Workflow of RecNet}
\label{sec:workflow}
\paratitle{Self-evolving Workflow.}
By integrating the forward propagation and backward optimization phases, we construct the complete workflow of RecNet, forming a continual and self-evolving structure.
From the perspective of a specific user or item, this workflow functions as a closed-loop process: 
Firstly, each client agent continuously receives relevant propagated preference information from other updated clients, which are temporarily stored in a local buffer. 
Then, during inference, the client selectively aggregates these information to form a refined profile that is then used for recommendation. 
After that, upon observing real user feedback, the client updates its own profile accordingly, and further performs backward optimization to adjust the propagation strategy. 
Finally, the client will propagate its updated preference information to other clients, forming a closed-loop cycle that enables RecNet to evolve continuously over time.

\paratitle{Efficient Implementation.} 
While proactive preference propagation offers benefits for recommendation, poorly designed propagation strategies can result in substantial computational overhead and inference latency.
To improve system efficiency, our RecNet introduces the following key advantages and enhancements:

$\bullet$ \textbf{Router-based Intermediaries}:
To determine both the content and the targets for propagation, an updated user or item client may need to directly communicate with many other clients, resulting in high communication complexity. In RecNet, the router-as-intermediaries architecture replaces direct point-to-point propagation with router-mediated propagation, where only a limited number of router agents are required to effectively manage preference propagation and achieve strong recommendation performance.

$\bullet$ \textbf{Asynchronous Optimization}:
To mitigate the additional overhead introduced by router updates, we adopt an asynchronous optimization strategy in the practical deployment of RecNet. 
Specifically, router-centric modules, such as router memories or quantities, are updated only after accumulating a batch of interaction signals from multiple clients. Client-centric modules, such as user/item profiles or filter memories, are updated directly when an interaction occurs. A detailed efficiency analysis can be found in Section~\ref{sec:efficiency}.

Taken together, the self-evolving workflow and the efficiency-oriented design enable RecNet to continuously propagate and refine user and item preferences with limited overhead. The detailed workflow is presented in Algorithm~\ref{alg:workflow}.




\begin{algorithm} 
	\caption{The Workflow of RecNet} 
    \renewcommand{\algorithmicrequire}{\textbf{Input:}}
	\renewcommand{\algorithmicensure}{\textbf{Output:}}
	\label{alg:workflow} 
	\begin{algorithmic}[1]
		\REQUIRE A set of users $\mathcal{U}$, a set of items $\mathcal{I}$ and a list of interaction record with negative candidates $\mathcal{D} = \{(u,i^+,i^-)\}$ 
		\ENSURE A fully constructed and operational RecNet system.
        \STATE Initialize updated user/item sets $\mathcal{S}_{update} \gets \emptyset$, the router set $\mathcal{R} = \{R_1, \dots, R_K\}$, the profile of users, items and routers $\mathcal{P}_u$,$\mathcal{P}_i$,$\mathcal{P}_r$, a set a filter memories $\mathcal{F}$ and a set of buffers $\mathcal{B} \gets \emptyset$ for each client.
        \WHILE{$\mathcal{D}$ is not empty}
            \STATE Fetch interaction record $d = (u, i^+, i^-)$
            \STATE \textbf{Stage 1: Start client-centric updation.}
            \STATE // Sample-based forward propagation
            \STATE $P^{pro}_c \gets \text{Prompt}_{merge}(\{R_c\},F_c, P_c), c \in \{u, i^+, i^-\}$
            \STATE $i_o \gets \text{Prompt}_{predict}(P_u^{pro}, P_{i+}^{pro}, P_{i-}^{pro})$
            \STATE // Sample-based backward optimization
            \STATE $L_M, \frac{\partial L_M}{\partial M} \gets \text{Prompt}_{gradient}(M, L)$
            \STATE $P_c^*,F_c^* \gets \text{Prompt}_{optimizer}(P_c, F_c, \frac{\partial L}{\partial P_c}, \frac{\partial L}{\partial F_c},)$
            \STATE $\mathcal{S}^*_{update} \gets \mathcal{S}_{update} + \{u, i^+, i^-\}$
            \STATE \textbf{Stage 2: Start router-centric updation.}
            \IF{$len(\mathcal{S}^*_{update}) \ge update\_size$ }
            \STATE // Batch-based backward optimization
            \STATE $\mathcal{R}^* \gets \text{Prompt}_{optimizer}(\mathcal{R}, \frac{\partial L}{\partial \mathcal{R}})$
            \STATE // Batch-based forward propagation
            \STATE $\mathcal{A} \gets \text{Prompt}_{extract}(P_c), c \in \mathcal{S}^*_{update}$
            \STATE $P_R^* = \text{Prompt}_{summarize}(\{a\}, P_R), a \in \mathcal{A}_R$
            \STATE $\text{Score}(R, c) \gets \text{sim}(e_{P_R}, e_{P_c}) \times \mathbb{I}\{\mathcal{A}_R \cap \mathcal{A}_c \neq \emptyset\}$
            \STATE Sending $R$ to corresponding client's buffer $B_c$, $c \in \{c_n|\text{Score}(R,c_n) > \tau\}$
            \ENDIF
        \ENDWHILE
	\end{algorithmic} 
\end{algorithm}

\section{Experiment}
\label{sec:experiment}

\subsection{Experiment Setup}
\subsubsection{Dataset}
Following previous work, we conduct experiments on three text-intensive subsets of Amazon review dataset: "CDs and Vinyl", "Office Products", and "Musical Instruments". Each of these datasets comprises user review data spanning from May 1996 to October
2018. For data preprocessing, we first remove unpopular users and items with less than five interactions through five-core filtering. Then, we create a historical interaction sequence sorted by timestamp for each. Due to the high cost of LLM inference, we further sample subsets from these three datasets. The statistics are shown in Table~\ref{tab:dataset}.

\begin{table}[t]
\small
  \centering
  \caption{Statistics of the preprocessed datasets.}
  \label{tab:dataset}
  \resizebox{0.75\columnwidth}{!}{
    \begin{tabular}{lrrrr}
    \toprule
    \textbf{Datasets} & \textbf{\#Users} & \textbf{\#Items} & \textbf{\#Inters.} & \textbf{Sparsity} \\
    \midrule
    \textbf{CDs (full)} & 84,522	& 56,883 & 961,572	& 99.98\%  \\
    \quad --\textbf{Sample} & 100   & 613   & 800   & 98.69\% \\
    \midrule
    \textbf{Office (full)} & 81,628 &	23,537 &	768,512 &99.96\%  \\
    \quad --\textbf{Sample} & 100   & 674   & 800   & 98.81\% \\
    \midrule
    \textbf{Instrument (full)} & 24,773	& 9,923 &	206,152 &	99.92\%  \\
    \quad --\textbf{Sample} & 100   & 752   & 800   & 98.93\% \\
    \bottomrule
    \end{tabular}}%
  \label{tab:dataset}%
\vspace{-10pt}
\end{table}%

\begin{table*}[]
\centering
\captionsetup{skip=5pt}
\caption{The overall performance comparisons between different baseline methods and RecNet. The best and second-best results are highlighted in bold and underlined font, respectively. All improvements are significant~($p$ < 0.05)}
\label{tab:main_res}
\resizebox{0.75\textwidth}{!}{
\renewcommand\arraystretch{1}
\begin{tabular}{lccccccccc}
\toprule 
\multicolumn{1}{c}{\multirow{2}{*}{Methods}} & \multicolumn{3}{c}{CDs}                           & \multicolumn{3}{c}{Office}                        & \multicolumn{3}{c}{Instruments}                   \\
\cmidrule(l){2-4} \cmidrule(l){5-7} \cmidrule(l){8-10} 
\multicolumn{1}{c}{}                         & N@1            & N@5              & N@10             & N@1            & N@5              & N@10             & N@1            & N@5              & N@10             \\
\midrule
Pop                                          & 0.0600          & 0.1792          & 0.2433          & 0.0600          & 0.1480           & 0.2137          & 0.1100          & 0.2146          & 0.4275          \\
BPR-sample                                   & 0.1200          & 0.2397          & 0.3074          & 0.0500          & 0.2461          & 0.2938          & 0.1400          & 0.3178          & 0.3584          \\
BPR-full                                     & 0.1900          & 0.4032          & 0.4596          & 0.2000           & 0.3341          & 0.4812          & 0.1600          & 0.3622          & 0.4380           \\
SASRec-sample                                & 0.1100          & 0.2866          & 0.3419          & 0.0900          & 0.2675          & 0.3894          & 0.1800          & 0.4037          & 0.4785          \\
SASRec-full                                  & \textbf{0.2700} & \underline{ 0.4438}    & \underline{ 0.5357}    & 0.1900          & 0.3761          & 0.5228          & 0.2300          & 0.4519          & \underline{ 0.5886}    \\
\midrule
LLMRank                                      & 0.1133        & 0.3184          & 0.3718          & 0.16 00         & 0.3272          & 0.4832          & 0.2133        & 0.4278          & 0.5011          \\
AgentCF                                      & 0.1600          & 0.3760           & 0.4898          & 0.1900          & 0.3506          & 0.5261          & 0.2500          & 0.4524          & 0.5462          \\
AgentCF++                                    & 0.1700          & 0.4018          & 0.5119          & 0.1867        & 0.3688          & 0.5208          & 0.2300          & 0.4606          & 0.5597          \\
KGLA                                         & 0.2000           & 0.4196          & 0.5187          & \underline{ 0.2100}    & \underline{ 0.3845}    & \underline{ 0.5435}    & \underline{ 0.2600}    & \underline{ 0.4762}    & 0.5742          \\
\midrule
RecNet                                       & \underline{ 0.2533}  & \textbf{0.4894} & \textbf{0.5745} & \textbf{0.2400} & \textbf{0.4373} & \textbf{0.5933} & \textbf{0.3200} & \textbf{0.5558} & \textbf{0.6417} \\
\bottomrule
\end{tabular}
}
\end{table*}

\subsubsection{Baseline Models}
We employ the following baselines:

\noindent \textbf{(1) Traditional recommendation models}:

$\bullet$ \textbf{Pop} ranks candidates based on their popularity, which is measured by the number of interactions.

$\bullet$ \textbf{BPR}~\cite{bpr} leverages matrix factorization to learn the representations of users and items by optimizing the BPR loss.

$\bullet$ \textbf{SASRec}~\cite{sasrec} is the first sequential recommender based on the
unidirectional self-attention mechanism

\noindent \textbf{(2) LLM-based recommendation models}:

$\bullet$ \textbf{LLMRank}~\cite{llmrank} uses the ChatGPT as a zero-shot ranker, considering user sequential interaction histories as conditions.

$\bullet$ \textbf{AgentCF}~\cite{agentcf} simulate the interaction between user agents and item agents for better recommendation.

$\bullet$ \textbf{AgentCF++}~\cite{agentcf++} leverages a dual-layer memory architecture to address cross-domain recommendation propblems.

$\bullet$ \textbf{KGLA}~\cite{kgla} uses knowledge graphs to augment the item information for agentic recommendation.

\subsubsection{Evaluataion Settings}
To evaluate the performance, we take
NDCG@K as a metric, where K is set to 1, 5 and 10. We employ the leave-one-out strategy for
evaluation. Specifically, we consider the last item in each historical
interaction sequence as the ground-truth item. By adopting the
model as a ranker, we rank the target item alongside nine randomly
sampled items. To further reduce randomness, we conduct three
repetitions of all test instances and report the average results.

\vspace{-5pt}

\subsubsection{Implementation Details}
We select Deepseek-R1-Distill-Qwen-32B as the backbone model for our main experiments. 
The initial router number $K$ is selected within {5, 10, 15, 20, 40}, and the threshold $\tau$ for preference routing is selected within {0.7, 0.75, 0.8, 0.85, 0.9}.
To adapt AgentCF++ and KGLA in our settings, we adjust the codes of AgentCF++ and manually generate the knowledge graph that are not involved from the source of KGLA.
We implement all traditional sequential recommendation models based on RecBole~\cite{recbole}. To ensure fair comparison, we set the embedding dimension of all models to 128 and obtain the best performance through hyperparameter grid search.
All experiments were carried out on eight A100 GPUs, each with 80GB of VRAM. 

\vspace{-5pt}

\subsection{Overall Performance}

The overall results of RecNet compared with other baselines are shown in Table~\ref{tab:main_res}. We can find:

Our proposed RecNet consistently outperforms the baselines across most evaluation metrics in three datasets. Compared to LLMRank and AgentCF-based baselines, RecNet brings significangt improvements, demonstrating the effectiveness and advantages of modeling dynamic preference propagation process. This helps the system to more accurately modeling user or item profiles and predict the following interactions.

Moreover, our model exhibits superior or comparable performance to traditional recommendation models when trained on datasets of the same scale (sampled datasets) or even the full datasets, which demonstrates the generalization capability of our approach. By using a small sample of datasets for simulation, RecNet has captured the accurate profile of users and items, showing high data efficiency for modeling preference representations. 

\begin{table}[h]
\centering
\captionsetup{skip=5pt}
\caption{Ablation Study.}
\label{tab:other_method}
\resizebox{0.95\linewidth}{!}{
\renewcommand\arraystretch{0.95}
\begin{tabular}{lcccc}
\toprule
\multicolumn{1}{c}{\multirow{2}{*}{Variants}} & \multicolumn{2}{c}{CDs}           & \multicolumn{2}{c}{Office}        \\
\cmidrule(l){2-3} \cmidrule(l){4-5}
\multicolumn{1}{c}{}                         & N@5              & N@10             & N@5              & N@10             \\
\midrule
RecNet                                       & \textbf{0.4894} & \textbf{0.5745} & \textbf{0.4373} & \textbf{0.5933} \\
\emph{w/o} CPR                                   & 0.4317          & 0.5378          & 0.3984          & 0.5682          \\
\emph{w/o} PPR                               & 0.4562          & 0.5481          & 0.4107          & 0.5813          \\
\quad -- \emph{w/o} Buffer                               & 0.4677          & 0.5610         & 0.4164          & 0.5826          \\
\quad -- \emph{w/o} Filter Memory                               & 0.4635          & 0.5649          & 0.4238          & 0.5870          \\
\emph{w/o} FPO                                 & 0.4360          & 0.5424          & 0.4052          & 0.5732          \\
\quad -- \emph{w/o} Opt. Filter Memory                            & 0.4722          & 0.5693          & 0.4155          & 0.5834          \\
\quad -- \emph{w/o} Opt. Router                                & 0.4580          & 0.5608          & 0.4102          & 0.5795     \\
\bottomrule
\end{tabular}
}
\end{table}

Furthermore, most LLM-based ranking recommendation models trained on sampled datasets achieve comparable results to traditional recommendation models trained on full datasets.
This indicates that, by leveraging their inherent world knowledge and textualized representations of users and items, LLM-based methods can effectively jointly learn collaborative filtering signals and semantic information even from limited data, achieving performance on par with models trained on full-scale datasets.

\subsection{Ablation Studies}
To validate the effectiveness of each modules in RecNet, we conduct ablation studies on the CDs and Office datasets, as shown in Table~\ref{tab:other_method} 

(1) \underline{\emph{w/o CPR}} eliminates the centralized preference routing module. Instead, we use dense retrieval method to directly retrieve top-k related client agents and propagate the entire updated profile to them. Moreover, for propagation optimization, we only optimize the filter memory for each client agent. We can see that this variant performs worse that RecNet across all datasets, which indicates that using router agent to perceive, integrate and dissemination the preference information serves as a crucial step in RecNet.

(2) \underline{\emph{w/o PPR}} omits the personalized preference reception module for each client agent. Under this settings, we also omit the backpropagation optimization of these modules. The results indicate that,  without the introduction of this module, the performance of model is limited, demonstrating the effectiveness of personalized preference reception.
Also, we conduct ablation experiments on two variants of PPR module, including \underline{\emph{w/o Buffer}} and \underline{\emph{w/o Filter Memory}}. The results of them both indicate the effectiveness of designing these two modules in helping client agents selectively integrating propagated preference information.

(3) \underline{\emph{w/o FPO}} removes the feedback-driven preference propagation module. The phenomenon demonstrates that, by dynamically update the propagation policy based on real-time feedback from ground-truth interaction, the model can refine its propagation pipeline and optimize the quality and accuracy of preference propagation. Similarly, we conduct ablation experiments on two variants of \emph{w/o FPO}, including \underline{\emph{w/o Opt. Filter Memory}} and \underline{\emph{w/o Opt. Router}}, representing not optimizing filter memory and not optimizing router agents respectively. The results show that optimizing both modules involved in the feed-forward propagation pipelines bring improvements to final recommendation performance.

\subsection{Further Analysis}
\subsubsection{Performance Comparison \wrt Different Model Backbones}

\begin{figure}[h]
\centering
\captionsetup{skip=5pt}
\includegraphics[width=0.9\linewidth]{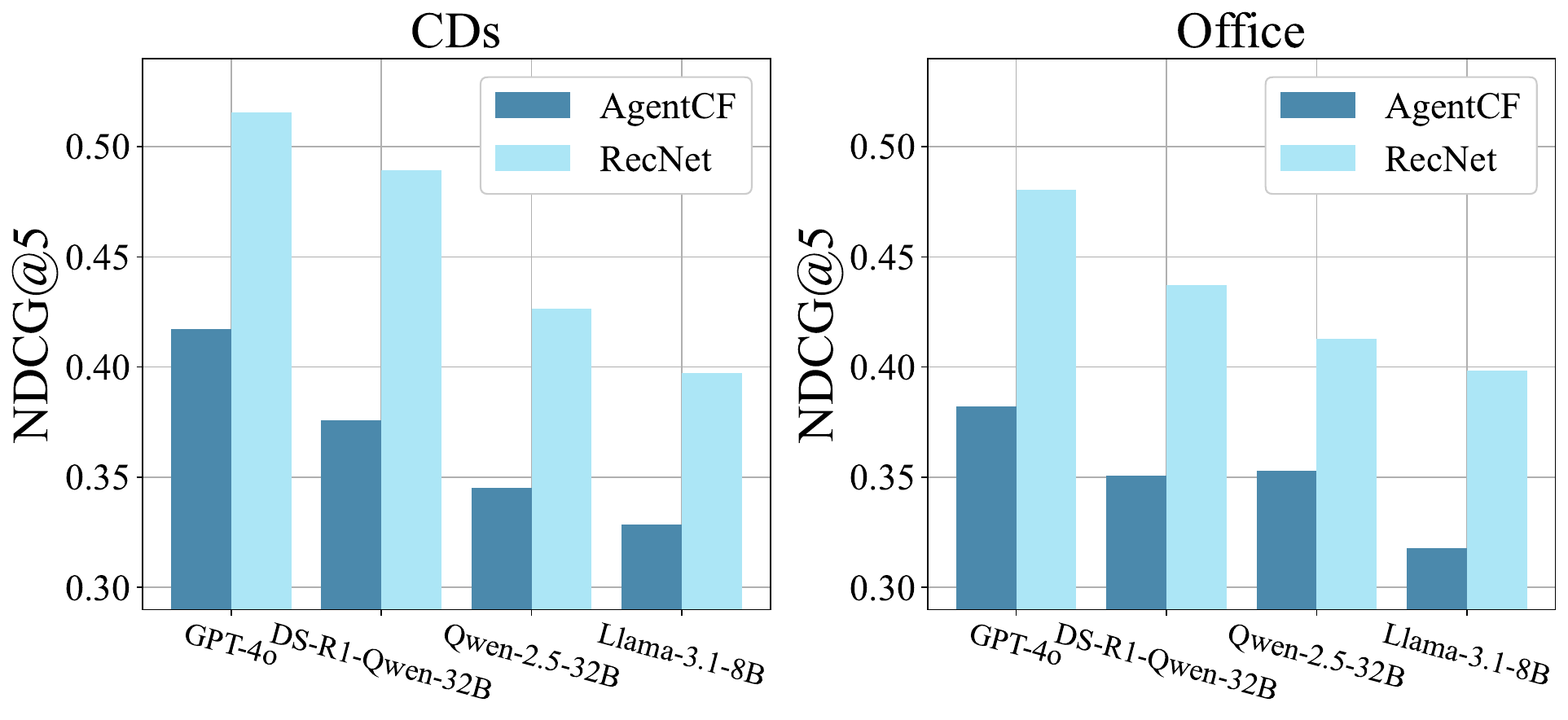}
\caption{Performance comparison \wrt different backbone models on CDs and Office.}
\label{fig:backbone}
\end{figure}

To evaluate the generalizability of RecNet across different model backbones, we conduct further experiments using different model backbones as agent modules in RecNet.
Specifically, we apply three open-source models~(Deepseek-R1-Distill-Qwen-32B, Qwen-2.5-32B, Llama-3.1-8B, respectively) and one closed-source model~(GPT-4o). The results are shown in Figure~\ref{fig:backbone}. We can observe that RecNet consistently outperforms other baselines across all tested backbones, demonstrating its strong generalizability to different model architectures. Notably, models with enhanced reasoning capabilities achieve higher performance than their counterparts without reasoning-oriented training, suggesting that the ability to perform fine-grained preference reasoning further amplifies the benefits of RecNet’s propagation mechanisms.

\begin{figure}[h]
\centering
\captionsetup{skip=5pt}
\includegraphics[width=0.87\linewidth]{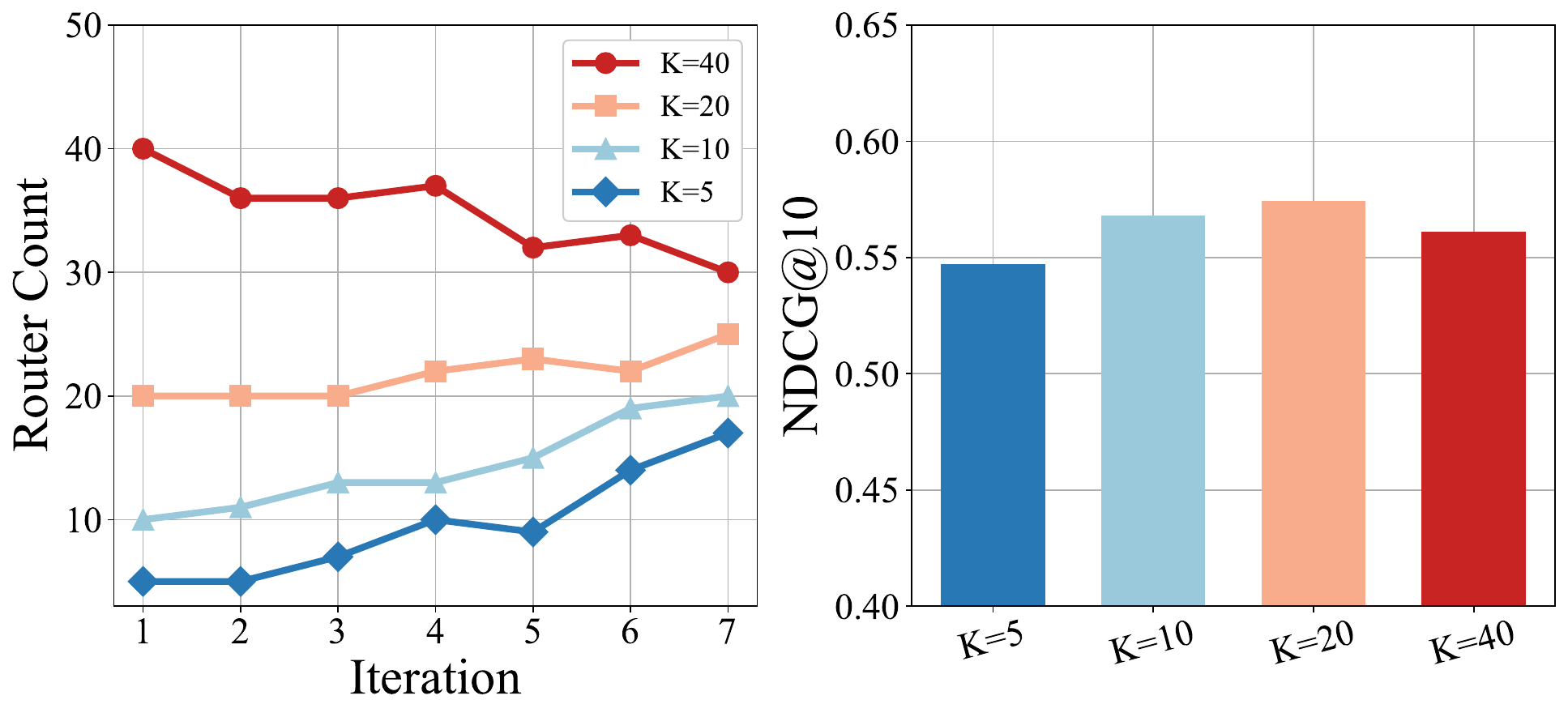}
\caption{The self-evolving phenomenon of feedback-driven propagation optimization \wrt different initial K.}
\label{fig:evolving}
\vspace{-10pt}
\end{figure}

\subsubsection{Evolution of RecNet \wrt Different Initial Router Numbers}
With Feedback-Driven Propagation Optimization, the attributes of router agents, including their quantities and router profiles, are dynamically updated. Hence, to demonstrate the evolution of router agents and to evaluate the generalizability of RecNet toward different initially set router numbers, we set the router number for CDs to 5, 10, 20, 40, respectively. The change of router numbers sampled from different steps and the final performance comparison are shown in Figure~\ref{fig:evolving}. We observe that setting the initial router number to 20 achieves the best recommendation performance. Notably, other initial settings also reach competitive results, demonstrating RecNet’s robustness to different initial router quantities. Regarding the evolution of router numbers, we can see that over time, the router counts from different initial settings gradually converge toward the optimal range. This indicates that feedback-driven propagation optimization can autonomously adjust the router quantities, enabling the system to self-evolve toward an optimal router distribution without manual intervention.
\vspace{-5pt}

\subsubsection{Performance Comparison \wrt cold-start users and items}

\begin{figure}[h]
\centering
\captionsetup{skip=5pt}
\includegraphics[width=0.9\linewidth]{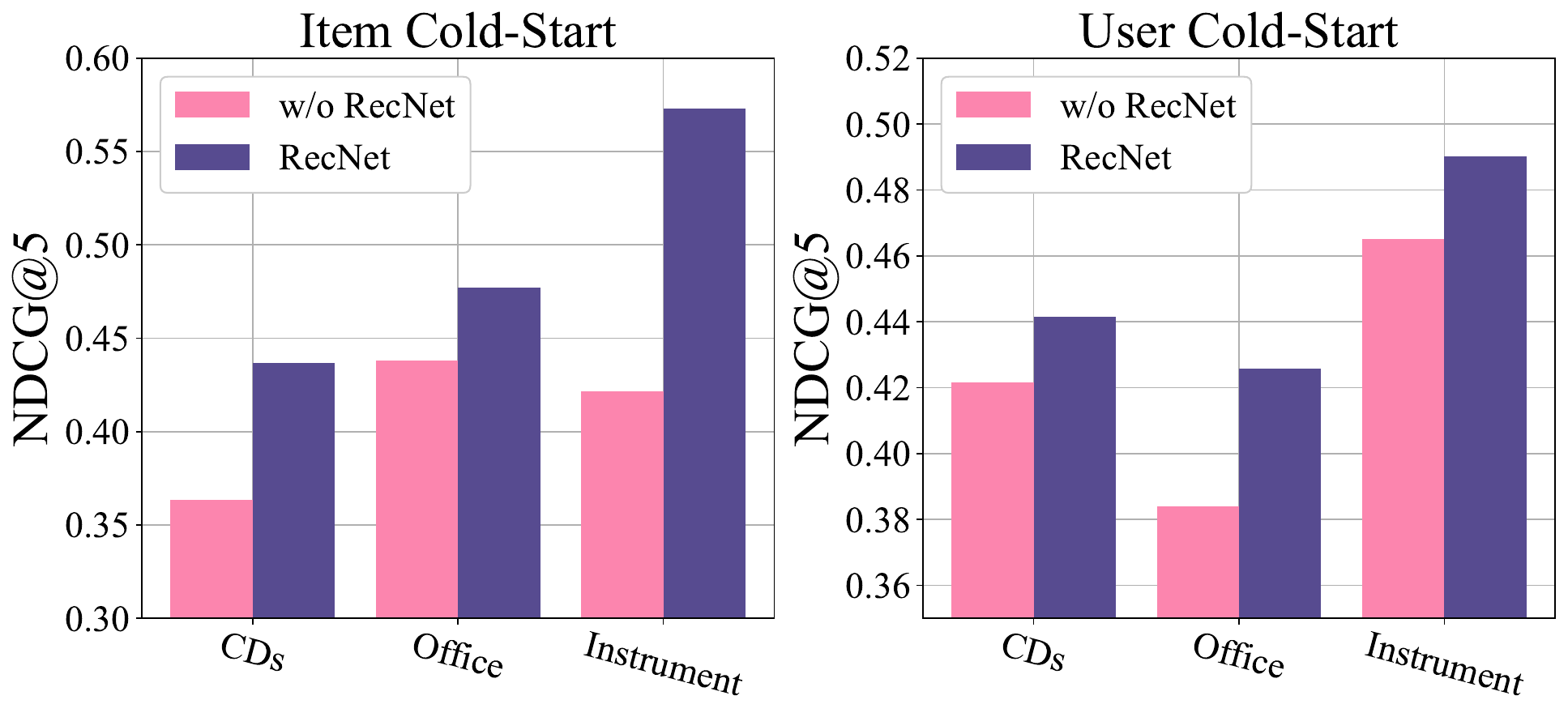}
\caption{Performance comparison \wrt RecNet-augmented cold-start items and cold-start users.}
\label{fig:cold-start}
\end{figure}

Traditional recommender systems face challenges in recommending new items or generating recommendations for users with sparse interaction histories, which are referred to as cold-start items and users. In this part, we aim to evaluate RecNet’s capability in handling cold-start users and items. 
For cold-start items, we sample those items involved in the test set but have not been interacted by users. We augment profiles of these cold-start items by connecting them to the routers and integrate the propagated router profiles into their own profiles immediately. Then, we add each of them with 9 randomly sampled items and prompt the user agent to select from the candidate item list. 
For cold-start users, we sample those users that only have one interaction during the evolution of RecNet. Similarly, we augment profiles of these cold-start users by connecting them to the routers and integrate the propagated router profiles into their own profiles immediately. 
The results in Figure~\ref{fig:cold-start} show that, compared with non-augmented cold-start item and user profiles, the profiles augmented by RecNet more accurately enrich and refine cold-start items and users, leading to improved recommendation performance.


\subsubsection{Different Propagation Methods}

\begin{table}[h]
\centering
\captionsetup{skip=5pt}
\caption{Performance comparison \wrt different propagation methods.}
\label{tab:propagation}
\resizebox{0.75\linewidth}{!}{
\renewcommand\arraystretch{0.95}
\begin{tabular}{lcccc}
\toprule
\multicolumn{1}{c}{\multirow{2}{*}{Methods}} & \multicolumn{2}{c}{CDs}           & \multicolumn{2}{c}{Office}        \\
\cmidrule(l){2-3} \cmidrule(l){4-5}
\multicolumn{1}{c}{}                         & N@5              & N@10             & N@5              & N@10             \\
\midrule
RecNet                                       & \textbf{0.4894} & \textbf{0.5745} & \textbf{0.4373} & \textbf{0.5933} \\
EM & 0.4317 &	0.5378 &	0.3984 &	0.5682 \\
EM+LR & 0.4560 &	0.5491 &	0.4037 &	0.5719 \\
EM+LR+LS & 0.4502 &	0.5552 &	0.4118 &	0.5782 \\
\bottomrule
\end{tabular}
}
\end{table}

To analyze the effectiveness of the client-router-client network architecture for preference propagation, we compare it with other basic methods that construct links and contents for propagation. From the perspective of link construction, we consider two variants:~(i) Embedding Retrieval~(EM), in which we directly retrieve top-k clients for an updated client profile. ~(ii) Embedding Retrieval + LLM Ranking(EM + LR), in which we first retrieve top-p clients(p > k), and then prompt LLMs to select k out of p clients for preference propagation. From the perspective of content propagation, we consider either to propagate the whole client profile, or to prompt LLMs to summarize the selected preference information and then propagate to target clients(LS). The results in Table~\ref{tab:propagation} show that, by perceiving and integrating fine-grained updated preference information as well as dynamic routing to highly relevant clients within the same community, RecNet achieves the highest performances compared to other counterparts.

\subsubsection{Efficiency Analysis}
\label{sec:efficiency}

\begin{table}[h]
\centering
\captionsetup{skip=5pt}
\caption{Efficiency analysis of different agentic recommender systems and preference propagation strategies.}
\label{tab:efficiency}
\resizebox{0.91\linewidth}{!}{
\renewcommand\arraystretch{0.95}
\begin{tabular}{lcccc}
\toprule
\multicolumn{1}{c}{\multirow{2}{*}{Methods}} & \multirow{2}{*}{Complexity} & \multirow{2}{*}{Time(min)}          & \multicolumn{2}{c}{Performance}        \\
 \cmidrule(l){4-5}
\multicolumn{1}{c}{}                         &              &             & N@5              & N@10             \\
\midrule
AgentCF & $\mathcal{O}(N)$ &	34 &	0.3760 &	0.4898 \\
RecNet \emph{w/o} router & $\mathcal{O}(\lambda N + NN')$ &	68 &	0.4936 &	0.5783 \\
RecNet \emph{w/o} asyn. & $\mathcal{O}(\lambda N + \mu M)$ &	49 &	0.4771 &	0.5716 \\
RecNet & $\mathcal{O}(\lambda N + M)$ &	41 &	0.4894 &	0.5745 \\
\bottomrule
\end{tabular}
}
\end{table}

For the efficiency analysis, we compare RecNet, AgentCF, and two RecNet variants that omit the efficiency-enhancing strategies described in Section~\ref{sec:workflow}. The comparison is performed along three dimensions: complexity (measured by the relative number of LLM calls per batch), time (total runtime on the CDs dataset), and recommendation performance.

Table~\ref{tab:efficiency} summarizes the results. AgentCF has a complexity of $\mathcal{O}(N)$, where $N$ denotes the number of updated users or items per batch. By incorporating preference propagation, RecNet achieves a complexity of $\mathcal{O}(\lambda N + M)$, where $M$ is the number of updated router agents per batch and $\lambda \approx 1.5$ in practice. Since $M \ll N$, this represents only a modest increase in computational complexity compared with AgentCF. In practice, while RecNet incurs slightly higher execution time due to additional LLM calls, it delivers substantial improvements in recommendation performance, demonstrating the effectiveness of preference propagation in enhancing the quality of recommendations.

The efficiency-enhancing strategies further mitigate the overhead introduced by preference propagation. Specifically, RecNet \emph{w/o} router requires extra LLM calls for each client to determine propagated content for downstream clients, resulting in an additional $NN'$ term in complexity, and RecNet \emph{w/o} asynchronous optimization triggers more frequent router updates, adding an extra $\mu M$ term where $\mu \gg 1$. In contrast, by leveraging the router-as-intermediaries architecture and asynchronous optimization, RecNet significantly reduces both LLM invocation and runtime costs, while maintaining strong recommendation performance, thereby achieving a favorable balance between efficiency and effectiveness.

\section{Related work}
\label{sec:related}

\paratitle{LLM-based Agents.}
LLM-based agents have gained extensive attention as a step toward artificial general intelligence~(AGI)~\cite{exploring, rise, agent4rec}. Unlike conventional static prompt–response models, they can autonomously reason, plan, and interact through natural language, forming dynamic decision-making loops~\cite{wang2024survey, zhao2023survey, park2023generative}. This enables decomposition of complex problems, continuous strategy refinement, and adaptation to diverse environments. A standard LLM-agent typically includes Profile, Memory, Planning, and Action modules~\cite{wang2024survey}, as exemplified by ReAct, Toolformer, and HuggingGPT~\cite{yao2023react, schick2023toolformer, shen2023hugginggpt}. Beyond single agents, multi-agent frameworks such as AutoGen, AgentPrune, and EvoMAC~\cite{hu2025agentgen, zhang2024cut, hu2024self} explore collaborative ecosystems where autonomous agents coordinate, share reasoning, and adaptively learn, enabling workflow-driven cooperation and human-like group dynamics. In RecNet, we aim at designing a preference propagation network based on the multi-agent recommender systems.

\paratitle{LLMs for Recommendation.}
As large language models (LLMs) have demonstrated remarkable
capabilities in text understanding and generation across various
domains~\cite{futuremedicine,cui2024chatlaw}, leveraging LLMs to enhance recommender systems
is gradually gaining widespread attention. Existing studies leveraged LLMs to generate auxiliary information to enhance the performance of traditional recommendation models~\cite{kar,llmrec,rlmrec}, or adapted LLMs to recommend items directly via zero-shot prompting or training~\cite{tallrec, lc-rec, llara, rosepo, sdpo}. Recently, some researchers attempt to incorporate agents into recommender systems, focusing on facilitating recommendations or user behavior simulations~\cite{agentcf, agentcf++, recmind, kgla, agent4rec}. 
However, to modeling representations of users or items, these studies mainly rely on interaction-based data to update the profile, which is noisy and partial to construct the whole recommender systems. Unlike these methods, we further modeling the preference propagation mechanism, which enables real-time and dynamic influence from users or items with updated profiles to other similar users and items.

\vspace{-5pt}
\section{Conclusion}
\label{sec:conclusion}

In this work, we presented RecNet, a self-evolving preference propagation framework designed to address the limitations of interaction-driven agentic recommender systems. By introducing a centralized preference routing mechanism, RecNet effectively coordinated communities of users and items, ensuring precise and efficient propagation of fine-grained preference updates. Complemented by the personalized preference reception module, user and item agents can selectively integrate propagated preferences in a personalized and context-aware manner. Furthermore, the feedback-driven propagation optimization enabled continuous self-evolution of the entire propagation pipeline by leveraging interaction feedback and simulating a multi-agent reinforcement learning framework with LLMs. Empirical evaluations demonstrated that RecNet captured dynamic preference evolution more accurately and adapted effectively to complex and evolving recommendation environments, offering a promising direction for next-generation recommender systems.

\bibliographystyle{ACM-Reference-Format}

\bibliography{ref}


\end{document}